\documentclass{midl} 

\usepackage{booktabs}
\usepackage{array}
\usepackage{colortbl}

\title[MedIL]{MedIL: Implicit Latent Spaces for Generating Heterogeneous Medical Images at Arbitrary Resolutions}

\midlauthor{
    \Name{Tyler Spears} \orcid{0000-0002-9104-9074} \Email{tas6hh@virginia.edu}\\
    \Name{Shen Zhu} \orcid{0009-0001-3978-0687} \Email{sz9jt@virginia.edu}\\
    \Name{Yinzhu Jin} \orcid{0009-0008-8904-446X} \Email{yj3cz@virginia.edu}\\
    \Name{Aman Shrivastava} \Email{as3ek@virginia.edu}\\
    \Name{P. Thomas Fletcher} \orcid{0000-0003-3417-2380}\Email{ptf8v@virginia.edu}\\
    \addr Dept. of Electrical \& Computer Engineering, University of Virginia, Charlottesville, VA, USA
}

\begin{document}

\maketitle

\begin{abstract}
In this work, we introduce MedIL, a first-of-its-kind autoencoder built for encoding medical images with heterogeneous sizes and resolutions for image generation. Medical images are often large and heterogeneous, where fine details are of vital clinical importance. Image properties change drastically when considering acquisition equipment, patient demographics, and pathology, making realistic medical image generation challenging. Recent work in latent diffusion models (LDMs) has shown success in generating images resampled to a fixed-size. However, this is a narrow subset of the resolutions native to image acquisition, and resampling discards fine anatomical details. MedIL utilizes implicit neural representations to treat images as continuous signals, where encoding and decoding can be performed at arbitrary resolutions without prior resampling. We quantitatively and qualitatively show how MedIL compresses and preserves clinically-relevant features over large multi-site, multi-resolution datasets of both T1w brain MRIs and lung CTs. We further demonstrate how MedIL can influence the quality of images generated with a diffusion model, and discuss how MedIL can enhance generative models to resemble raw clinical acquisitions.
\end{abstract}

\begin{keywords}
LDM, autoencoder, INR
\end{keywords}

\section{Introduction}
\label{sec:introduction}
Medical images come in all shapes and sizes. For example, a T1w magnetic resonance image (MRI) of the brain has a spatial resolution dependent upon scan sequence, magnetic field strength, etc. Similarly, lung computational tomography (CT) images may be acquired with different out-of-plane resolutions \cite{ArmatoIII.etal.2015}. These parameters further depend on the pathology of interest, the quality of scanner hardware, and other external factors. This diversity of image resolutions poses a challenge for generative models that aim to learn a distribution of medical images. If such models are to fully represent real-world clinical images, anatomical details must be preserved throughout the model.

Recent denoising diffusion probability models (DDPM) are more tractable when trained on a compressed latent space, with an accompanying autoencoder, in a latent diffusion model (LDM). This reliance on compression implies that the generative power of an LDM is only as good as its autoencoder. However, most LDM autoencoders require a fixed input size to produce a fixed latent space shape, which further requires input data to be resampled and manipulated. Thus, the limitations of these autoencoders require image homogenization, which narrows image diversity and hinders the DDPM from the start.

To resolve this bottleneck, we introduce MedIL - \textbf{Med}ical images from \textbf{I}mplicit \textbf{L}atent spaces. MedIL is a novel autoencoder that utilizes implicit neural representations (INRs) to flexibly encode medical images \textit{without any prior resampling}. The INRs used in MedIL consider images as discrete samples of 3D spatially-continuous signals with physical coordinates, which allows for sampling at any arbitrary coordinate. This flexibility enables MedIL to encode any resolution of medical images, produce latent spaces of any size, and decode latent space samples to any arbitrary resolution. Our contributions are as follows:
\begin{itemize}
    \item We propose and describe MedIL, a novel autoencoder built for encoding and decoding medical images with heterogeneous resolutions.
    \item We demonstrate MedIL as an autoencoder for T1w brain MRIs and lung CTs.
    \item We show how MedIL's flexibility improves downstream generative image tasks.
    \item We publicly release our MedIL implementation which includes model implementations and utility functions for future work in spatially-continuous autoencoders. (\url{https://github.com/TylerSpears/medil})
\end{itemize}

\section{Background}

\subsection{Related Work}
\label{subsec:related_work}
Deep generative models for medical images have been applied to a variety of imaging tasks for several years. More recently, these models have been improved with the denoising diffusion probabilistic model (DDPM) \cite{Ho.etal.2020b}. Authors in \cite{Dorjsembe.etal.2022} were one of the first to demonstrate unconditional generation of 3D T1w brain MRIs using a DDPM, with subsequent work \cite{Pinaya.etal.2022a} improving image quality. The more specific problem of arbitrary-resolution image generation is a sub-field of image generation where the source and/or target distributions do not have a fixed size. Many arbitrary-resolution works use generative adversarial networks (GANs) enhanced with positional encoding. INR-GAN was proposed by \cite{Skorokhodov.etal.2021}, which used a hypernetwork to produce the weights of a per-image INR. Authors in \cite{Ntavelis.etal.2022} utilized positional encodings and scale-consistency to generate partial images. Anyres-GAN \cite{Chai.etal.2022a} generated nature images of varying aspect ratios using a GAN with pixel coordinates. There are few works in this sub-field with DDPMs, and to the best of our knowledge, there are no prior works on arbitrary-resolution medical image generation with DDPMs.

\subsection{Latent Diffusion Models}
\label{sec:ldm}
Traditional diffusion models denoise Gaussian samples in pixel space, requiring intensive computation for high-resolution images and sequential sampling steps. Latent Diffusion Models (LDMs) \cite{Rombach.etal.2022b} address this by operating in compressed latent space. First, an autoencoder perceptually compresses inputs into a lower-dimensional latent space. Next, a diffusion model is trained in this latent representation, enabling scalable high-resolution synthesis. Building on standard diffusion models \cite{Ho.etal.2020b}, LDMs \cite{Rombach.etal.2022b} implement diffusion in compressed latent space. The forward process gradually adds Gaussian noise via $q(x_t|x_{t-1}) = \mathcal{N}(\sqrt{1-\beta_t}x_{t-1}, \beta_t\mathbf{I})$ over $T$ steps, converging to $\mathcal{N}(0,\mathbf{I})$. The reverse process learns to denoise through $p_\theta(x_{t-1}|x_t) = \mathcal{N}(\boldsymbol{\mu}_\theta(x_t,t), \boldsymbol{\Sigma}_\theta(x_t,t))$, trained using a hybrid loss \cite{Nichol.Dhariwal.2021a} (where $\epsilon_\theta$ predicts the added noise):
\begin{equation}
    L_{\text{hybrid}} = \underbrace{\mathbb{E}_{t,x_0,\epsilon} \|\epsilon - \epsilon_\theta(x_t,t)\|^2}_{L_{\text{simple}}} + \lambda \underbrace{L_{\text{vlb}}}_{\text{VLB objective}}
\end{equation}

\subsection{Continuous Image Representations with INRs}
\label{sec:inr}

INRs are MLPs that learn a continuously-valued representation of a discretely-sampled signal. While many INRs learn to represent a single image or scene, such as neural radiance fields (NeRF) and sinusoidal representation networks (SIREN) \cite{Mildenhall.etal.2021,Sitzmann.etal.2020}, other methods consider INRs as a learned interpolation kernel over space. For example, the Local Implicit Image Function (LIIF) is an arbitrary-size super-resolution method for upsampling images to any zoom factor \cite{Chen.etal.2021}.

For notational consistency, let us define $\Omega \subset \mathbb{R}^d$ as a continuous, bounded coordinate space which contains the coordinates of all images in our dataset. Also, denote variables $C \subset \Omega$ as discrete, rectilinear grids of coordinate vectors $c \in C$ (thus, $c \in \mathbb{R}^d$). Then, LIIF considers every pixel $x$ in image $X$ as a discrete sample with an associated coordinate $c_X \in C_X$, where $d=2$ for 2D images\footnote{We constrain our notation to 2D in describing previous work, but our MedIL implementation is in 3D.}. The output of LIIF is determined by a query coordinate $c_Q \in C_Q$ that can be calculated by a transformation of $c_X$, e.g. zoom. LIIF uses an MLP $f_\theta$ to create a function $F \left(f_\theta, X, C_X, C_Q \right)$ that reconstructs the image $X$ at coordinates $C_Q$.

To avoid grid artifacts at pixel boundaries, LIIF performs $4$ forward-passes at each nearest-neighbor point (the local ``ensemble'') in $C_X$ relative to every $c_Q$. Denote the coordinate $c_X^{i,j}$ with $i, j \in \{0, 1\}$, as the coordinate in $C_X$ that is the $(i,j)$th nearest pixel to a $c_Q$, and let $X[c_X]$ be a sampling operation of $X$ at $c_X$. Then, the INR forward passes are linearly combined as 
\begin{equation}
\label{eq:liif_ensemble}
F \left( f_\theta, X, C_X, c_Q \right) = 
    \sum_i \sum_j 
        w^{i,j}
        f_\theta \left(X\left[c_X^{i,j}\right], c_X^{i,j}, c_Q \right),
\end{equation}
where $w^{i,j}$ is the linear interpolation weight of the $i,j$'th forward pass.

Subsequent INR works used positional encodings to reduce the spectral bias in INRs \cite{Tancik.etal.2020,Xu.etal.2022}. The local texture estimator (LTE) \cite{Lee.Jin.2022}, used in MedIL, builds on LIIF by learning these positional encodings as frequencies and phase offsets for image super-resolution.

\section{MedIL}
\label{sec:medil}

\begin{figure}[t]
\begin{center}
\includegraphics[width=\linewidth]{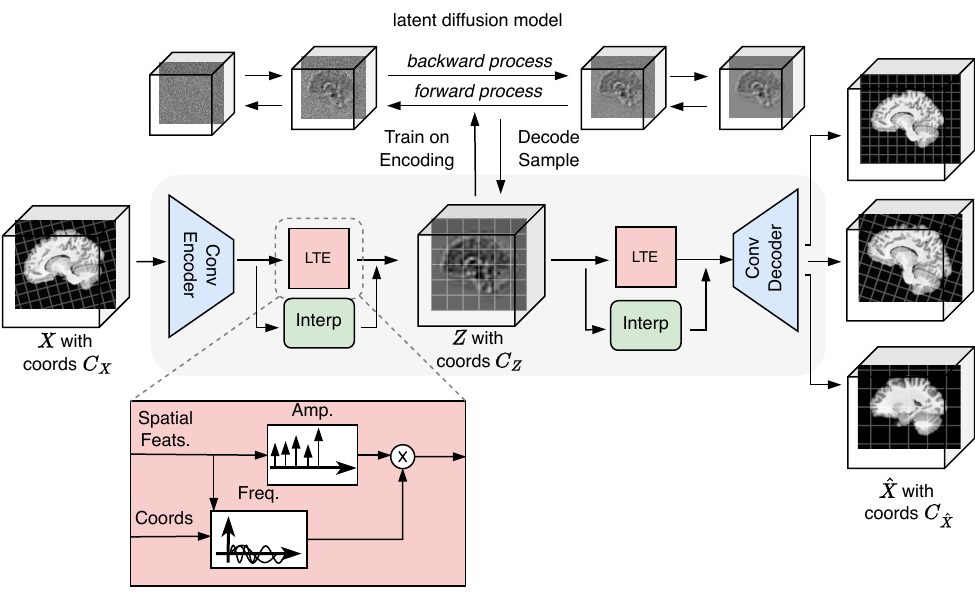}
\end{center}
   \caption{MedIL architecture. Input volumes $X$ are encoded into latent volumes $Z$, which can then be decoded to arbitrary output resolutions or orientations.}
\label{fig:medil}
\end{figure}

MedIL is an autoencoder architecture where all volumes are treated as a sample of a spatially-continuous signal using a mix of convolutional and INR modules in both the encoder and decoder. The MedIL architecture is illustrated in Figure \ref{fig:medil}. The input, latent, and output volumes ($X$, $Z$, and $\hat{X}$, respectively) all have associated coordinate grids $C_X$, $C_Z$, and $C_{\hat{X}}$ which are contained in the bounded, continuous coordinate space $\Omega$. Medical images are unique in that voxel locations have defined physical coordinates, so $\Omega$ may be defined by the scanner's field of view (FoV) or a template space \cite{Evans.etal.1993}. We then consider encoding as a transformation of $X$ into the $Z$ via their coordinate grids $C_X$ and $C_Z$. Similarly, decoding is a transformation of $Z$ to $\hat{X}$ using a coordinate transformation from $C_Z$ to $C_{\hat{X}}$. As illustrated in Figure \ref{fig:medil}, MedIL is capable of decoding (and encoding) images of arbitrary resolution and orientation so long as their coordinates are contained within $\Omega$. This flexibility allow MedIL to encode fixed-size, but feature-rich, latent representations for subsequent image generation tasks.

The MedIL encoder comprises a fully convolutional backbone and an LTE network. This backbone encodes spatial context for the sampling module and spatially compresses $X$ by some integer factor(s). This helps ``spread out'' the compression burden between layers, and reduces the memory requirements for a 3D LTE. The LTE module then continuously resamples the convolutional backbone's output to the coordinates $C_Z$, producing $Z$. The decoder must then resample $Z$ into $\hat{X}$ via $C_Z$ and $C_{\hat{X}}$. The decoder mirrors the encoder, starting with an LTE module to perform the ``variable'' resampling, and a convolutional backbone to upsample by an integer factor.

\textbf{Training with Large Volumes.} Due to the large size of raw medical images (i.e. $512^3$ voxels), it is infeasible to train MedIL over entire volumes. So, based on work in \cite{Chai.etal.2022a}, we break up training MedIL into two stages. First, we pretrain MedIL on volumes downsampled to a standard size such that features across the entire FoV of $\Omega$ are available to MedIL. Second, patches of full-resolution volumes are randomly sampled from a tissue mask. These $\hat{X}$ volumes have a fixed \textit{voxel} shape, but their \textit{physical} resolutions may vary. We also ensure spatial context around $\Hat{X}$ by selecting $C_Z$ such that the boundary points of $C_Z$ encompass $C_{\Hat{X}}$ with some buffer space. We similarly ``exscribe'' $C_X$ around $C_Z$.

\section{Experiments \& Results}

\subsection{Data}
\textbf{T1w Brain MRIs.} To maximize the diversity of acquisition sites and image parameters, we chose to train MedIL on T1w brain MRIs from five different publicly available datasets: the Young Adult Human Connectome Project (HCP) \cite{VanEssen.etal.2013}, the Autism Brain Imaging Data Exchange (ABIDE) 1 and 2 \cite{DiMartino.etal.2014,DiMartino.etal.2017a}, the Open Access Series of Imaging Studies-3 (OASIS3) dataset \cite{LaMontagne.etal.2019}, and the Center of Reproducible Research (CoRR) brain dataset \cite{Zuo.etal.2014}. We selected healthy adult ($\ge$ 21 years of age) subjects with a total of roughly 3,700 unique subjects and slightly over 7,000 volumes, as various datasets provided multiple scans per subject.

\textbf{Lung CTs.} We further validated MedIL on a second image modality, CT images of lungs from the Lung Image Database Consortium and Image Database Resource Initiative (LIDC-IDRI) \cite{ArmatoIII.etal.2015}. The LIDC-IDRI dataset consists of 1,000 lung CTs with out-of-plane resolutions ranging between 0.7mm to 5.0mm. Details are given in Appendix \ref{append:preproc}, but we emphasize that no spatial interpolation was performed during preprocessing.

\subsection{Comparison Models}
\label{subsec:comparison_models}
We sought to compare MedIL to previously established fully-convolutional autoencoders found in the medical image literature. For the T1w brain MRIs, we recreated the LDM (both autoencoder and diffusion model) architecture and training parameters from \cite{Pinaya.etal.2022a}. We similarly followed \cite{Khader.etal.2023b} for constructing and training an LDM for chest CTs. These models, which we name as Conv. LDMs, are built for fixed-size images, and we aim to reconstruct volumes at their native resolutions. So, we must first train these models on fixed-size volumes, then interpolate their output to the image's native resolution. We resampled all T1w brain MRIs to be AC-PC aligned, with a 1mm isotropic resolution and size $160 \times 224 \times 160$. We also resampled all lung CTs to a resolution of 2.5mm isotropic with the shape $128^3$. These resampled data formed the training set for both the Conv. LDMs and the MedIL pretraining stage. We also compared MedIL to an ablation variant with the LTE module stripped out and replaced with only trilinear interpolation, which we name MedIL-Interp.

\subsection{Experiment 1. Autoencoder Reconstruction of T1w Brain MRIs}

\begin{figure}[t]
\begin{center}
    \includegraphics[width=\linewidth]{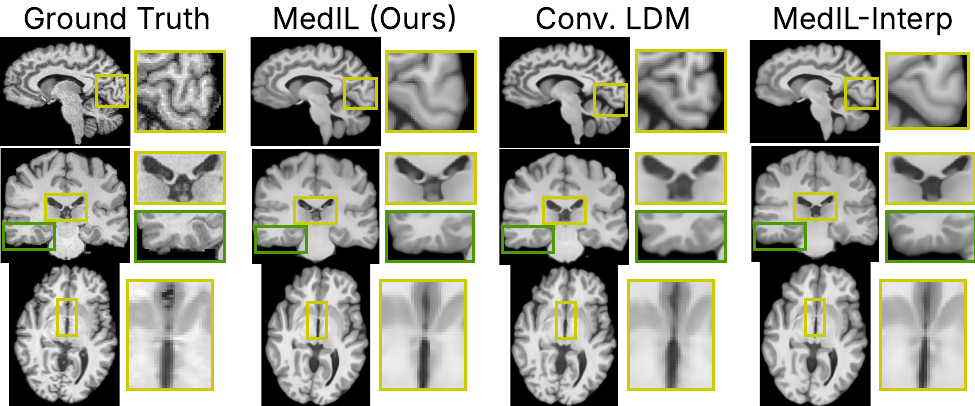}
\end{center}
   \caption{Reconstruction of real, native-space T1w brain MRIs.}
    \label{fig:brain_recon}
\end{figure}

\begin{table}
    \caption{Results for reconstructing native-space T1w Brain MRIs (Experiment 1) and native-space lung CTs (Experiment 2). All images were scaled to the range $[0,1]$.}
    \label{table:recon_results}

\centering
\begin{tabular}{l|lllll}
 &
  \cellcolor[HTML]{EFEFEF}$\downarrow$ $\ell_1$ &
  $\uparrow$PSNR &
  \cellcolor[HTML]{EFEFEF}$\uparrow$SSIM &
  $\uparrow$MS-SSIM &
  \cellcolor[HTML]{EFEFEF}$\downarrow$Perceptual \\ \hline
\textit{Experiment 1: Brain MRI} &                                           &           &                                  &          &                                  \\ \hline
Conv. LDM &
  \cellcolor[HTML]{EFEFEF}\textbf{0.035424} &
  \textbf{30.188066} &
  \cellcolor[HTML]{EFEFEF}\textbf{0.879466} &
  \textbf{0.975497} &
  \cellcolor[HTML]{EFEFEF}\textbf{0.000183} \\
MedIL-Interp (Ablation)          & \cellcolor[HTML]{EFEFEF}0.041470          & 29.222708 & \cellcolor[HTML]{EFEFEF}0.846165 & 0.959142 & \cellcolor[HTML]{EFEFEF}0.000375 \\
MedIL (Ours)                     & \cellcolor[HTML]{EFEFEF}0.037323          & 30.040161 & \cellcolor[HTML]{EFEFEF}0.868112 & 0.969519 & \cellcolor[HTML]{EFEFEF}0.000302 \\ \hline
\textit{Experiment 2: Lung CT}   &                                           &           &                                  &          &                                  \\ \hline
Conv. LDM                        & \cellcolor[HTML]{EFEFEF}\textbf{0.026827} & 32.547900 & \cellcolor[HTML]{EFEFEF}0.827005 & 0.955557 & \cellcolor[HTML]{EFEFEF}0.003036 \\
MedIL-Interp (Ablation)          & \cellcolor[HTML]{EFEFEF}0.030542          & 31.815599 & \cellcolor[HTML]{EFEFEF}0.833598 & 0.949818 & \cellcolor[HTML]{EFEFEF}0.000913 \\
MedIL (Ours) &
  \cellcolor[HTML]{EFEFEF}0.027418 &
  \textbf{32.770102} &
  \cellcolor[HTML]{EFEFEF}\textbf{0.851286} &
  \textbf{0.959340} &
  \cellcolor[HTML]{EFEFEF}\textbf{0.000779}
\end{tabular}

\end{table}

In our first experiment, we evaluated MedIL's reconstruction performance for native-resolution T1w brain MRIs. We compared MedIL to a Conv. LDM \cite{Pinaya.etal.2022a} with trilinear resampling into the native space. The latent space consisted of 3 channels, and its coordinate grid $C_Z$ was AC-PC aligned with 8mm spacing. Test sets consisted of 50 randomly selected subjects per dataset, giving 250 test volumes. Model reconstructions were evaluated by $\ell_1$, peak signal-to-noise ratio (PSNR), the structural similarity index measure (SSIM) \cite{Wang.etal.2004}, multi-scale SSIM \cite{Wang.etal.2003b}, and the Med3D learned perceptual metric \cite{Chen.etal.2019b}.

Reconstructions examples for Experiment 1 are shown in Figure \ref{fig:brain_recon}, and quantitative results are shown in Table \ref{table:recon_results}. MedIL produces high-quality reconstructions of native-resolution T1w brain MRIs across multiple datasets. Even with resolutions as high as $0.7$mm, greater than a $10\times$ downsampling to $Z$, MedIL preserves small details. Row 1 highlights how MedIL can maintain the separation of gyri, whereas Conv. LDMs may struggle (though the opposite may often occur). The gold boxes in Figure \ref{fig:brain_recon} rows 2 and 3 show how fine anatomical features are lost in the resampling required for Conv. LDMs, such as the boundary between the fornix and the Thalamus. In row 3, the inferior edge of the anterior commissure is disconnected for Conv. LDM, but maintained by MedIL. These results also demonstrate some weaknesses of MedIL, such as a less clear separation between white and gray matter. Additionally, we observed that at higher resolutions, MedIL may insert extraneous gyri as shown in the green boxes in row 2. We hypothesize this is an effect of the pretraining on 1mm data. For the quantitative results in Table \ref{table:recon_results}, MedIL is highly competitive with Conv. LDM in all tested metrics. MedIL also outperformed the ablation model in all given metrics, demonstrating the advantage of the learned resampling modules.

\subsection{Experiment 2. Autoencoder Reconstruction of Lung CTs}

\begin{figure}[t]
\begin{center}
    \includegraphics[width=\linewidth]{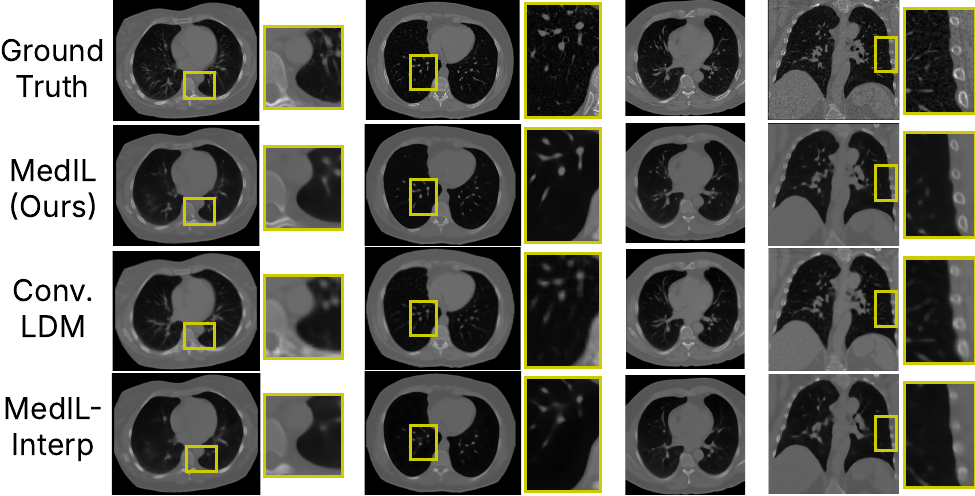}
\end{center}
   \caption{Reconstruction of real, native-space lung CTs. 
   }
    \label{fig:lung_recon}
\end{figure}

In Experiment 2, we further validated MedIL as a general medical image autoencoder via reconstructing lung CTs at native resolutions. Unlike T1w brain MRIs, with resolutions between $0.7$ to $1.3$mm, the out-of-plane resolutions of lung CTs span $0.6$ and $5$mm for the LIDC-IDRI dataset. Furthermore, lung CTs may be extremely large (up to $512^3$ voxels), posing a computational challenge when resampling cannot be performed. MedIL for lung CT was compared to the Conv. LDM described in \cite{Khader.etal.2023b}. Following this previous work, Conv. LDM was trained on CTs from the LIDC-IDRI dataset resampled to $2.5$mm isotropic ($128$ voxels). Evaluation was performed over a test set of 99 random subjects in the CT's native resolution, with Conv. LDM reconstructions trilinearly resampled to the native resolution. Test performance metrics are the same as Experiment 1.

Example lung CT reconstructions are shown in Figure \ref{fig:lung_recon}, and quantitative metrics are shown in Table \ref{table:recon_results}. MedIL effectively reconstructs different tissue and bone boundaries over a range of CT resolutions. For example, in the left and right columns of Figure \ref{fig:lung_recon}, MedIL maintains a clear border between tissue and air and even layers within bone. This is opposed to Conv. LDM where the necessary trilinear upsampling has blurred all boundaries. Inside the lungs, MedIL reconstructs the larger-scale nodules with clarity, but struggles with small, narrow branches, a clear example of the spectral bias problem \cite{Tancik.etal.2020}. Conv. LDM blurs and weakens these same branches, along with the nodules, but are not entirely erased. Despite this, MedIL quantitatively outperforms the comparison models in all metrics besides $\ell_1$. This performance is likely due to MedIL's preservation of tissue boundaries, and each metric's lower weighting of small, low-intensity bronchial features.

\subsection{Experiment 3. Generating T1w Brain MRIs at Different Resolutions}

\begin{figure}[t]
\begin{center}
    \includegraphics[width=\linewidth]{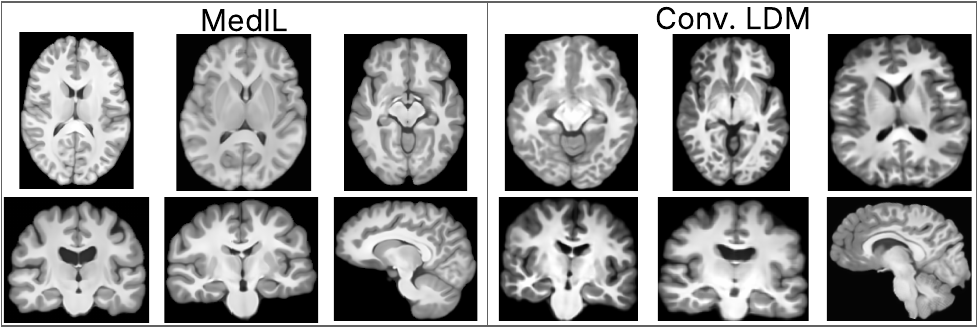}
\end{center}
   \caption{Example T1w MRIs generated with a DDPM on latent space samples.}
    \label{fig:brain_samples}
\end{figure}

\begin{table}
    \caption{Generative model results for Experiment 3 trained with encodings from Experiments 1 and 2.}
    \label{table:gen_metrics}

    \centering
    \begin{tabular}{lllllll}
     &
       &
       &
      \multicolumn{2}{c}{$\uparrow$ Coverage} &
      \multicolumn{2}{c}{$\uparrow$ Density} \\
    \multicolumn{1}{l|}{} &
      \cellcolor[HTML]{EFEFEF}$\downarrow$ FID &
      $\downarrow$ MS-SSIM &
      \cellcolor[HTML]{EFEFEF}$k=10$ &
      $k=30$ &
      \cellcolor[HTML]{EFEFEF}$k=10$ &
      $k=30$ \\ \hline
    \multicolumn{1}{l|}{\textit{Brain MRI}} &
       &
       &
       &
       &
       &
       \\ \hline
    \multicolumn{1}{l|}{Conv. LDM} &
      \cellcolor[HTML]{EFEFEF}0.144 &
      \textbf{0.6903} &
      \cellcolor[HTML]{EFEFEF}0.556 &
      0.856 &
      \cellcolor[HTML]{EFEFEF}0.131 &
      0.1657 \\
    \multicolumn{1}{l|}{MedIL (Ours)} &
      \cellcolor[HTML]{EFEFEF}\textbf{0.035} &
      0.8090 &
      \cellcolor[HTML]{EFEFEF}\textbf{0.675} &
      \textbf{0.888} &
      \cellcolor[HTML]{EFEFEF}\textbf{0.2185} &
      \textbf{0.2857} \\
    \multicolumn{1}{l|}{Real Data} &
      \cellcolor[HTML]{EFEFEF}- &
      0.6801 &
      \cellcolor[HTML]{EFEFEF}- &
      - &
      \cellcolor[HTML]{EFEFEF}- &
      - \\ \hline
    \multicolumn{1}{l|}{\textit{Lung CT}} &
       &
       &
       &
       &
       &
       \\ \hline
    \multicolumn{1}{l|}{Conv. LDM} &
      \cellcolor[HTML]{EFEFEF}\textbf{0.00769} &
      0.4798 &
      \cellcolor[HTML]{EFEFEF}\textbf{0.902} &
      \textbf{0.997} &
      \cellcolor[HTML]{EFEFEF}\textbf{0.323} &
      \textbf{0.394} \\
    \multicolumn{1}{l|}{MedIL (Ours)} &
      \cellcolor[HTML]{EFEFEF}0.01410 &
      \textbf{0.4547} &
      \cellcolor[HTML]{EFEFEF}0.706 &
      0.971 &
      \cellcolor[HTML]{EFEFEF}0.249 &
      0.362 \\
    \multicolumn{1}{l|}{Real Data} &
      \cellcolor[HTML]{EFEFEF}- &
      0.4313 &
      \cellcolor[HTML]{EFEFEF}- &
      - &
      \cellcolor[HTML]{EFEFEF}- &
      -
    \end{tabular}

\end{table}

In our final experiment, we sought to demonstrate how MedIL latent encodings may influence downstream image generation. Using the MedIL and Conv. LDM models from Experiments 1 and 2, we encoded the entire training and validation sets into latent samples and trained a UNet \cite{Ronneberger.etal.2015d} DDPM for 1,000 epochs. We sampled 1,000 latent volumes from the DDPM, and each model decoded its samples into 1mm AC-PC aligned space for the brain MRIs, and $0.7\times0.7\times2.5$mm space for the lung CTs. For brain MRI, no resampling was required for Conv. LDM, and MedIL decoded samples into the 1mm output space with its decoder LTE. Whereas lung CT required resampling of the Conv. LDM output using trilinear interpolation. We compared the synthetic and real data distributions multiple metrics including the Frechet Inception distance (FID) on Med3D \cite{Chen.etal.2019b}, coverage, density \cite{Naeem.etal.2020}, and an MS-SSIM diversity metric. We also wanted to visualize synthetic T1w MRIs in native space. To emulate native spaces, we fit a Gaussian mixture model with 4 components to the resolutions and orientations found in the real data. We then sampled this transformation distribution for each DDPM sample, which became the output spaces. Finally, the DDPM samples were decoded into their respective output spaces.

The quantitative results of decoding samples to a fixed space are found in Table \ref{table:gen_metrics}, and examples of the native-space decodings of brain MRIs are shown in Figure \ref{fig:brain_samples}. For brain MRI, samples trained on MedIL encodings more closely match the real data distribution when compared to Conv. LDM on FID, coverage, and density, but had lower diversity according to MS-SSIM over $1,000$ random pairs within each dataset. We hypothesize that this performance advantage may be due to MedIL's sharper tissue boundaries (see Figure \ref{fig:brain_samples}). In native-space decodings, qualitatively, we found that Conv. LDM often distorted small-scale anatomical features. In Figure \ref{fig:brain_samples}, the top-left Conv. LDM sample shows a warped cerebral aqueduct, and the bottom-center shows a missing (or attenuated) fornix. This was not universal for Conv. LDM, but we did not see such distortions with MedIL. However, similar to the Experiment 1 results, MedIL did struggle to decode some samples into higher resolutions (roughly 0.8mm), producing unrealistic gyral topology. Overall, MedIL samples can be effectively decoded to many resolutions.

For generated lung CTs, we find that Conv. LDM more closely matches the real data distribution according to FID, coverage, and density, but MedIL has slightly better diversity according to MS-SSIM (over 200 random pairings within dataset). We hypothesize that, despite the higher reconstruction performance of MedIL in lung CTs, the features of importance in the Med3D network, which was used to estimate the feature distributions of each dataset, may prioritize bronchial tubes over bone, muscle, etc. These features would not be given preference by MS-SSIM.

\subsection{Discussion}
In this work, we introduced MedIL, a novel autoencoder architecture built for encoding medical images of all shapes and sizes. We have demonstrated MedIL's capabilities for reconstructing medical images with different native spaces, for both T1w MRI and lung CT, and how MedIL's flexible representations can improve fixed-size convolutional LDMs. However, this work is only an introduction of MedIL. Future work should focus on lowering the spectral bias found in many INRs, and expanding to conditional image generation to push towards generating clinical images with fine-scale, anatomically correct features.

\clearpage  




\midlacknowledgments{We thank Dr. Miaomiao Zhang for insightful feedback and generously shared computational resources.}

\bibliography{references}

\begin{thebibliography}{32}
\providecommand{\natexlab}[1]{#1}
\providecommand{\url}[1]{\texttt{#1}}
\expandafter\ifx\csname urlstyle\endcsname\relax
  \providecommand{\doi}[1]{doi: #1}\else
  \providecommand{\doi}{doi: \begingroup \urlstyle{rm}\Url}\fi

\bibitem[Ardekani and Bachman(2009)]{Ardekani.Bachman.2009a}
Babak~A. Ardekani and Alvin~H. Bachman.
\newblock Model-based automatic detection of the anterior and posterior commissures on {{MRI}} scans.
\newblock \emph{NeuroImage}, 46\penalty0 (3):\penalty0 677--682, July 2009.
\newblock ISSN 1053-8119.

\bibitem[Armato~III et~al.(2015)Armato~III, McLennan, Bidaut, {McNitt-Gray}, Meyer, Reeves, Zhao, Aberle, Henschke, Hoffman, Kazerooni, MacMahon, Van~Beek, Yankelevitz, Biancardi, Bland, Brown, Engelmann, Laderach, Max, Pais, Qing, Roberts, Smith, Starkey, Batra, Caligiuri, Farooqi, Gladish, Jude, Munden, Petkovska, Quint, Schwartz, Sundaram, Dodd, Fenimore, Gur, Petrick, Freymann, Kirby, Hughes, Casteele, Gupte, Sallam, Heath, Kuhn, Dharaiya, Burns, Fryd, Salganicoff, Anand, Shreter, Vastagh, Croft, and Clarke]{ArmatoIII.etal.2015}
Samuel~G. Armato~III, Geoffrey McLennan, Luc Bidaut, Michael~F. {McNitt-Gray}, Charles~R. Meyer, Anthony~P. Reeves, Binsheng Zhao, Denise~R. Aberle, Claudia~I. Henschke, Eric~A. Hoffman, Ella~A. Kazerooni, Heber MacMahon, Edwin~J.R. Van~Beek, David Yankelevitz, Alberto~M. Biancardi, Peyton~H. Bland, Matthew~S. Brown, Roger~M. Engelmann, Gary~E. Laderach, Daniel Max, Richard~C. Pais, David~P.Y. Qing, Rachael~Y. Roberts, Amanda~R. Smith, Adam Starkey, Poonam Batra, Philip Caligiuri, Ali Farooqi, Gregory~W. Gladish, C.~Matilda Jude, Reginald~F. Munden, Iva Petkovska, Leslie~E. Quint, Lawrence~H. Schwartz, Baskaran Sundaram, Lori~E. Dodd, Charles Fenimore, David Gur, Nicholas Petrick, John Freymann, Justin Kirby, Brian Hughes, Alessi~Vande Casteele, Sangeeta Gupte, Maha Sallam, Michael~D. Heath, Michael~H. Kuhn, Ekta Dharaiya, Richard Burns, David~S. Fryd, Marcos Salganicoff, Vikram Anand, Uri Shreter, Stephen Vastagh, Barbara~Y. Croft, and Laurence~P. Clarke.
\newblock Data {{From LIDC-IDRI}}, 2015.

\bibitem[Chai et~al.(2022)Chai, Gharbi, Shechtman, Isola, and Zhang]{Chai.etal.2022a}
Lucy Chai, Micha{\"e}l Gharbi, Eli Shechtman, Phillip Isola, and Richard Zhang.
\newblock Any-{{Resolution Training}} for~{{High-Resolution Image Synthesis}}.
\newblock In Shai Avidan, Gabriel Brostow, Moustapha Ciss{\'e}, Giovanni~Maria Farinella, and Tal Hassner, editors, \emph{Computer {{Vision}} -- {{ECCV}} 2022}, pages 170--188, Cham, 2022. Springer Nature Switzerland.
\newblock ISBN 978-3-031-19787-1.

\bibitem[Chen et~al.(2019)Chen, Ma, and Zheng]{Chen.etal.2019b}
Sihong Chen, Kai Ma, and Yefeng Zheng.
\newblock {{Med3D}}: {{Transfer Learning}} for {{3D Medical Image Analysis}}, July 2019.

\bibitem[Chen et~al.(2021)Chen, Liu, and Wang]{Chen.etal.2021}
Yinbo Chen, Sifei Liu, and Xiaolong Wang.
\newblock Learning {{Continuous Image Representation}} with {{Local Implicit Image Function}}, April 2021.

\bibitem[Dale et~al.(1999)Dale, Fischl, and Sereno]{Dale.etal.1999}
Anders~M. Dale, Bruce Fischl, and Martin~I. Sereno.
\newblock Cortical {{Surface-Based Analysis}}: {{I}}. {{Segmentation}} and {{Surface Reconstruction}}.
\newblock \emph{NeuroImage}, 9\penalty0 (2):\penalty0 179--194, February 1999.
\newblock ISSN 1053-8119.

\bibitem[Di~Martino et~al.(2014)Di~Martino, Yan, Li, Denio, Castellanos, Alaerts, Anderson, Assaf, Bookheimer, Dapretto, Deen, Delmonte, Dinstein, {Ertl-Wagner}, Fair, Gallagher, Kennedy, Keown, Keysers, Lainhart, Lord, Luna, Menon, Minshew, Monk, Mueller, M{\"u}ller, Nebel, Nigg, O'Hearn, Pelphrey, Peltier, Rudie, Sunaert, Thioux, Tyszka, Uddin, Verhoeven, Wenderoth, Wiggins, Mostofsky, and Milham]{DiMartino.etal.2014}
A.~Di~Martino, C.-G. Yan, Q.~Li, E.~Denio, F.~X. Castellanos, K.~Alaerts, J.~S. Anderson, M.~Assaf, S.~Y. Bookheimer, M.~Dapretto, B.~Deen, S.~Delmonte, I.~Dinstein, B.~{Ertl-Wagner}, D.~A. Fair, L.~Gallagher, D.~P. Kennedy, C.~L. Keown, C.~Keysers, J.~E. Lainhart, C.~Lord, B.~Luna, V.~Menon, N.~J. Minshew, C.~S. Monk, S.~Mueller, R.-A. M{\"u}ller, M.~B. Nebel, J.~T. Nigg, K.~O'Hearn, K.~A. Pelphrey, S.~J. Peltier, J.~D. Rudie, S.~Sunaert, M.~Thioux, J.~M. Tyszka, L.~Q. Uddin, J.~S. Verhoeven, N.~Wenderoth, J.~L. Wiggins, S.~H. Mostofsky, and M.~P. Milham.
\newblock The autism brain imaging data exchange: Towards a large-scale evaluation of the intrinsic brain architecture in autism.
\newblock \emph{Mol Psychiatry}, 19\penalty0 (6):\penalty0 659--667, June 2014.
\newblock ISSN 1476-5578.

\bibitem[Di~Martino et~al.(2017)Di~Martino, O'Connor, Chen, Alaerts, Anderson, Assaf, Balsters, Baxter, Beggiato, Bernaerts, Blanken, Bookheimer, Braden, Byrge, Castellanos, Dapretto, Delorme, Fair, Fishman, Fitzgerald, Gallagher, Keehn, Kennedy, Lainhart, Luna, Mostofsky, M{\"u}ller, Nebel, Nigg, O'Hearn, Solomon, Toro, Vaidya, Wenderoth, White, Craddock, Lord, Leventhal, and Milham]{DiMartino.etal.2017a}
Adriana Di~Martino, David O'Connor, Bosi Chen, Kaat Alaerts, Jeffrey~S. Anderson, Michal Assaf, Joshua~H. Balsters, Leslie Baxter, Anita Beggiato, Sylvie Bernaerts, Laura M.~E. Blanken, Susan~Y. Bookheimer, B.~Blair Braden, Lisa Byrge, F.~Xavier Castellanos, Mirella Dapretto, Richard Delorme, Damien~A. Fair, Inna Fishman, Jacqueline Fitzgerald, Louise Gallagher, R.~Joanne~Jao Keehn, Daniel~P. Kennedy, Janet~E. Lainhart, Beatriz Luna, Stewart~H. Mostofsky, Ralph-Axel M{\"u}ller, Mary~Beth Nebel, Joel~T. Nigg, Kirsten O'Hearn, Marjorie Solomon, Roberto Toro, Chandan~J. Vaidya, Nicole Wenderoth, Tonya White, R.~Cameron Craddock, Catherine Lord, Bennett Leventhal, and Michael~P. Milham.
\newblock Enhancing studies of the connectome in autism using the autism brain imaging data exchange {{II}}.
\newblock \emph{Sci Data}, 4\penalty0 (1):\penalty0 170010, March 2017.
\newblock ISSN 2052-4463.

\bibitem[Dorjsembe et~al.(2022)Dorjsembe, Odonchimed, and Xiao]{Dorjsembe.etal.2022}
Zolnamar Dorjsembe, Sodtavilan Odonchimed, and Furen Xiao.
\newblock Three-{{Dimensional Medical Image Synthesis}} with {{Denoising Diffusion Probabilistic Models}}.
\newblock In \emph{Medical {{Imaging}} with {{Deep Learning}}}, April 2022.

\bibitem[Evans et~al.(1993)Evans, Collins, Mills, Brown, Kelly, and Peters]{Evans.etal.1993}
A.C. Evans, D.L. Collins, S.R. Mills, E.D. Brown, R.L. Kelly, and T.M. Peters.
\newblock {{3D}} statistical neuroanatomical models from 305 {{MRI}} volumes.
\newblock In \emph{1993 {{IEEE Conference Record Nuclear Science Symposium}} and {{Medical Imaging Conference}}}, pages 1813--1817 vol.3, October 1993.

\bibitem[Ho et~al.(2020)Ho, Jain, and Abbeel]{Ho.etal.2020b}
Jonathan Ho, Ajay Jain, and Pieter Abbeel.
\newblock Denoising {{Diffusion Probabilistic Models}}, December 2020.

\bibitem[Khader et~al.(2023)Khader, {M{\"u}ller-Franzes}, Tayebi~Arasteh, Han, Haarburger, {Schulze-Hagen}, Schad, Engelhardt, Bae{\ss}ler, Foersch, Stegmaier, Kuhl, Nebelung, Kather, and Truhn]{Khader.etal.2023b}
Firas Khader, Gustav {M{\"u}ller-Franzes}, Soroosh Tayebi~Arasteh, Tianyu Han, Christoph Haarburger, Maximilian {Schulze-Hagen}, Philipp Schad, Sandy Engelhardt, Bettina Bae{\ss}ler, Sebastian Foersch, Johannes Stegmaier, Christiane Kuhl, Sven Nebelung, Jakob~Nikolas Kather, and Daniel Truhn.
\newblock Denoising diffusion probabilistic models for {{3D}} medical image generation.
\newblock \emph{Sci Rep}, 13\penalty0 (1):\penalty0 7303, May 2023.
\newblock ISSN 2045-2322.

\bibitem[LaMontagne et~al.(2019)LaMontagne, Benzinger, Morris, Keefe, Hornbeck, Xiong, Grant, Hassenstab, Moulder, Vlassenko, Raichle, Cruchaga, and Marcus]{LaMontagne.etal.2019}
Pamela~J. LaMontagne, Tammie~LS Benzinger, John~C. Morris, Sarah Keefe, Russ Hornbeck, Chengjie Xiong, Elizabeth Grant, Jason Hassenstab, Krista Moulder, Andrei~G. Vlassenko, Marcus~E. Raichle, Carlos Cruchaga, and Daniel Marcus.
\newblock {{OASIS-3}}: {{Longitudinal Neuroimaging}}, {{Clinical}}, and {{Cognitive Dataset}} for {{Normal Aging}} and {{Alzheimer Disease}}, December 2019.
\newblock ISSN 1901-4902.

\bibitem[Lee and Jin(2022)]{Lee.Jin.2022}
Jaewon Lee and Kyong~Hwan Jin.
\newblock Local {{Texture Estimator}} for {{Implicit Representation Function}}.
\newblock In \emph{Proceedings of the {{IEEE}}/{{CVF Conference}} on {{Computer Vision}} and {{Pattern Recognition}}}, pages 1929--1938, 2022.

\bibitem[Loshchilov and Hutter(2019)]{Loshchilov.Hutter.2019}
Ilya Loshchilov and Frank Hutter.
\newblock Decoupled {{Weight Decay Regularization}}, January 2019.

\bibitem[Mildenhall et~al.(2021)Mildenhall, Srinivasan, Tancik, Barron, Ramamoorthi, and Ng]{Mildenhall.etal.2021}
Ben Mildenhall, Pratul~P. Srinivasan, Matthew Tancik, Jonathan~T. Barron, Ravi Ramamoorthi, and Ren Ng.
\newblock {{NeRF}}: Representing scenes as neural radiance fields for view synthesis.
\newblock \emph{Commun. ACM}, 65\penalty0 (1):\penalty0 99--106, December 2021.
\newblock ISSN 0001-0782.

\bibitem[Naeem et~al.(2020)Naeem, Oh, Uh, Choi, and Yoo]{Naeem.etal.2020}
Muhammad~Ferjad Naeem, Seong~Joon Oh, Youngjung Uh, Yunjey Choi, and Jaejun Yoo.
\newblock Reliable {{Fidelity}} and {{Diversity Metrics}} for {{Generative Models}}.
\newblock In \emph{Proceedings of the 37th {{International Conference}} on {{Machine Learning}}}, pages 7176--7185. PMLR, November 2020.

\bibitem[Nichol and Dhariwal(2021)]{Nichol.Dhariwal.2021a}
Alex Nichol and Prafulla Dhariwal.
\newblock Improved {{Denoising Diffusion Probabilistic Models}}, February 2021.

\bibitem[Ntavelis et~al.(2022)Ntavelis, Shahbazi, Kastanis, Timofte, Danelljan, and Van~Gool]{Ntavelis.etal.2022}
Evangelos Ntavelis, Mohamad Shahbazi, Iason Kastanis, Radu Timofte, Martin Danelljan, and Luc Van~Gool.
\newblock Arbitrary-{{Scale Image Synthesis}}.
\newblock In \emph{2022 {{IEEE}}/{{CVF Conference}} on {{Computer Vision}} and {{Pattern Recognition}} ({{CVPR}})}, pages 11523--11532, New Orleans, LA, USA, June 2022. IEEE.
\newblock ISBN 978-1-6654-6946-3.

\bibitem[Pinaya et~al.(2022)Pinaya, Tudosiu, Dafflon, Da~Costa, Fernandez, Nachev, Ourselin, and Cardoso]{Pinaya.etal.2022a}
Walter H.~L. Pinaya, Petru-Daniel Tudosiu, Jessica Dafflon, Pedro~F. Da~Costa, Virginia Fernandez, Parashkev Nachev, Sebastien Ourselin, and M.~Jorge Cardoso.
\newblock Brain {{Imaging Generation}} with~{{Latent Diffusion Models}}.
\newblock In Anirban Mukhopadhyay, Ilkay Oksuz, Sandy Engelhardt, Dajiang Zhu, and Yixuan Yuan, editors, \emph{Deep {{Generative Models}}}, pages 117--126, Cham, 2022. Springer Nature Switzerland.
\newblock ISBN 978-3-031-18576-2.

\bibitem[Rombach et~al.(2022)Rombach, Blattmann, Lorenz, Esser, and Ommer]{Rombach.etal.2022b}
Robin Rombach, Andreas Blattmann, Dominik Lorenz, Patrick Esser, and Bj{\"o}rn Ommer.
\newblock High-{{Resolution Image Synthesis With Latent Diffusion Models}}.
\newblock In \emph{Proceedings of the {{IEEE}}/{{CVF Conference}} on {{Computer Vision}} and {{Pattern Recognition}}}, pages 10684--10695, 2022.

\bibitem[Ronneberger et~al.(2015)Ronneberger, Fischer, and Brox]{Ronneberger.etal.2015d}
Olaf Ronneberger, Philipp Fischer, and Thomas Brox.
\newblock U-{{Net}}: {{Convolutional Networks}} for {{Biomedical Image Segmentation}}, May 2015.

\bibitem[S{\'e}gonne et~al.(2004)S{\'e}gonne, Dale, Busa, Glessner, Salat, Hahn, and Fischl]{Segonne.etal.2004}
F.~S{\'e}gonne, A.~M. Dale, E.~Busa, M.~Glessner, D.~Salat, H.~K. Hahn, and B.~Fischl.
\newblock A hybrid approach to the skull stripping problem in {{MRI}}.
\newblock \emph{NeuroImage}, 22\penalty0 (3):\penalty0 1060--1075, July 2004.
\newblock ISSN 1053-8119.

\bibitem[Sitzmann et~al.(2020)Sitzmann, Martel, Bergman, Lindell, and Wetzstein]{Sitzmann.etal.2020}
Vincent Sitzmann, Julien N.~P. Martel, Alexander~W. Bergman, David~B. Lindell, and Gordon Wetzstein.
\newblock Implicit {{Neural Representations}} with {{Periodic Activation Functions}}, June 2020.

\bibitem[Skorokhodov et~al.(2021)Skorokhodov, Ignatyev, and Elhoseiny]{Skorokhodov.etal.2021}
Ivan Skorokhodov, Savva Ignatyev, and Mohamed Elhoseiny.
\newblock Adversarial {{Generation}} of {{Continuous Images}}.
\newblock In \emph{Proceedings of the {{IEEE}}/{{CVF Conference}} on {{Computer Vision}} and {{Pattern Recognition}}}, pages 10753--10764, 2021.

\bibitem[Tancik et~al.(2020)Tancik, Srinivasan, Mildenhall, {Fridovich-Keil}, Raghavan, Singhal, Ramamoorthi, Barron, and Ng]{Tancik.etal.2020}
Matthew Tancik, Pratul~P. Srinivasan, Ben Mildenhall, Sara {Fridovich-Keil}, Nithin Raghavan, Utkarsh Singhal, Ravi Ramamoorthi, Jonathan~T. Barron, and Ren Ng.
\newblock Fourier {{Features Let Networks Learn High Frequency Functions}} in {{Low Dimensional Domains}}, June 2020.

\bibitem[Tustison et~al.(2010)Tustison, Avants, Cook, Zheng, Egan, Yushkevich, and Gee]{Tustison.etal.2010a}
Nicholas~J. Tustison, Brian~B. Avants, Philip~A. Cook, Yuanjie Zheng, Alexander Egan, Paul~A. Yushkevich, and James~C. Gee.
\newblock {{N4ITK}}: {{Improved N3 Bias Correction}}.
\newblock \emph{IEEE Transactions on Medical Imaging}, 29\penalty0 (6):\penalty0 1310--1320, June 2010.
\newblock ISSN 1558-254X.

\bibitem[Van~Essen et~al.(2013)Van~Essen, Smith, Barch, Behrens, Yacoub, and Ugurbil]{VanEssen.etal.2013}
David~C. Van~Essen, Stephen~M. Smith, Deanna~M. Barch, Timothy E.~J. Behrens, Essa Yacoub, and Kamil Ugurbil.
\newblock The {{WU-Minn Human Connectome Project}}: {{An}} overview.
\newblock \emph{NeuroImage}, 80:\penalty0 62--79, October 2013.
\newblock ISSN 1053-8119.

\bibitem[Wang et~al.(2003)Wang, Simoncelli, and Bovik]{Wang.etal.2003b}
Z.~Wang, E.P. Simoncelli, and A.C. Bovik.
\newblock Multiscale structural similarity for image quality assessment.
\newblock In \emph{The {{Thrity-Seventh Asilomar Conference}} on {{Signals}}, {{Systems}} \& {{Computers}}, 2003}, volume~2, pages 1398--1402 Vol.2, November 2003.

\bibitem[Wang et~al.(2004)Wang, Bovik, Sheikh, and Simoncelli]{Wang.etal.2004}
Zhou Wang, A.C. Bovik, H.R. Sheikh, and E.P. Simoncelli.
\newblock Image quality assessment: From error visibility to structural similarity.
\newblock \emph{IEEE TIP}, 13\penalty0 (4):\penalty0 600--612, April 2004.
\newblock ISSN 1941-0042.

\bibitem[Xu et~al.(2022)Xu, Wang, and Shi]{Xu.etal.2022}
Xingqian Xu, Zhangyang Wang, and Humphrey Shi.
\newblock {{UltraSR}}: {{Spatial Encoding}} is a {{Missing Key}} for {{Implicit Image Function-based Arbitrary-Scale Super-Resolution}}, July 2022.

\bibitem[Zuo et~al.(2014)Zuo, Anderson, Bellec, Birn, Biswal, Blautzik, Breitner, Buckner, Calhoun, Castellanos, Chen, Chen, Chen, Chen, Colcombe, Courtney, Craddock, Di~Martino, Dong, Fu, Gong, Gorgolewski, Han, He, He, Ho, Holmes, Hou, Huckins, Jiang, Jiang, Kelley, Kelly, King, LaConte, Lainhart, Lei, Li, Li, Li, Lin, Liu, Liu, Liu, Liu, Lu, Lu, Luna, Luo, Lurie, Mao, Margulies, Mayer, Meindl, Meyerand, Nan, Nielsen, O'Connor, Paulsen, Prabhakaran, Qi, Qiu, Shao, Shehzad, Tang, Villringer, Wang, Wang, Wei, Wei, Weng, Wu, Xu, Yang, Yang, Zang, Zhang, Zhang, Zhang, Zhang, Zhao, Zhen, Zhou, Zhu, and Milham]{Zuo.etal.2014}
Xi-Nian Zuo, Jeffrey~S. Anderson, Pierre Bellec, Rasmus~M. Birn, Bharat~B. Biswal, Janusch Blautzik, John C.~S. Breitner, Randy~L. Buckner, Vince~D. Calhoun, F.~Xavier Castellanos, Antao Chen, Bing Chen, Jiangtao Chen, Xu~Chen, Stanley~J. Colcombe, William Courtney, R.~Cameron Craddock, Adriana Di~Martino, Hao-Ming Dong, Xiaolan Fu, Qiyong Gong, Krzysztof~J. Gorgolewski, Ying Han, Ye~He, Yong He, Erica Ho, Avram Holmes, Xiao-Hui Hou, Jeremy Huckins, Tianzi Jiang, Yi~Jiang, William Kelley, Clare Kelly, Margaret King, Stephen~M. LaConte, Janet~E. Lainhart, Xu~Lei, Hui-Jie Li, Kaiming Li, Kuncheng Li, Qixiang Lin, Dongqiang Liu, Jia Liu, Xun Liu, Yijun Liu, Guangming Lu, Jie Lu, Beatriz Luna, Jing Luo, Daniel Lurie, Ying Mao, Daniel~S. Margulies, Andrew~R. Mayer, Thomas Meindl, Mary~E. Meyerand, Weizhi Nan, Jared~A. Nielsen, David O'Connor, David Paulsen, Vivek Prabhakaran, Zhigang Qi, Jiang Qiu, Chunhong Shao, Zarrar Shehzad, Weijun Tang, Arno Villringer, Huiling Wang, Kai Wang, Dongtao Wei, Gao-Xia Wei, Xu-Chu
  Weng, Xuehai Wu, Ting Xu, Ning Yang, Zhi Yang, Yu-Feng Zang, Lei Zhang, Qinglin Zhang, Zhe Zhang, Zhiqiang Zhang, Ke~Zhao, Zonglei Zhen, Yuan Zhou, Xing-Ting Zhu, and Michael~P. Milham.
\newblock An open science resource for establishing reliability and reproducibility in functional connectomics.
\newblock \emph{Sci Data}, 1\penalty0 (1):\penalty0 140049, December 2014.
\newblock ISSN 2052-4463.

\end{thebibliography}

\appendix

\section{Experiment Details}
For all experiments, one epoch corresponded to 200 batches of samples (whole volumes or patches), with an effective batch size of 8.

\subsection{Data Preprocessing}
\label{append:preproc}

\textbf{T1w Brain MRI.} Preprocessing of the T1w brain MRIs was performed on volumes in such a way as to focus the analysis on brain structures without compromising the high-frequency properties unique to each scan. The T1w scans were bias-corrected using N4ITK \cite{Tustison.etal.2010a} and skull-stripped with Freesurfer \cite{Dale.etal.1999,Segonne.etal.2004}. All brain MRIs were then aligned to a common orientation to account for head position in the scanner, which was performed using landmark detection of the anterior commissure-posterior commissure (AC-PC) plane with the \texttt{acpcdetect} tool \cite{Ardekani.Bachman.2009a}. We note that we only transformed the coordinate space such that each volume would be aligned in $\Omega$, there was \textit{no} resampling performed and all image content was kept its native resolution and orientation. Scans from HCP were excluded from this preprocessing as they were already preprocessed with specialized techniques at a high spatial resolution (0.7mm isotropic). Finally, each volume was scaled to a range of $0.0$ to $1.0$ such that the volume's 99th percentile would have a value of $1.0$.

\textbf{Lung CT.} Preprocessing of lung CTs followed the procedure described in \cite{Khader.etal.2023b} except that CT volumes were only cropped or padded to a common FoV rather than resampled. Transformations between $X$ and the latent coordinates $C_Z$ consisted only of scaling and translation with no rotation. Tissue masks were created by thresholding at -200 Hounsfield units (HUs), along with a morphological fill operation to include air in the lungs. Intensities were scaled such that the range $-1,000$ to $2,000$ HUs was mapped to $[0.0, 1.0]$.

\subsection{Network Architectures}
\textbf{Experiment 1.} MedIL's convolutional backbone consisted of 4 residual blocks, all with kernel size 3, 2 residual units per block, and 2 subunits per residual unit. Channel sizes were 32, 64, 96, 96, and instance norm was used throughout the convolutional backbone. Downscaling factors were 2, 2, 1, with downsampling being performed with convolutions with strides $> 1$. The convolutional layers in the decoder mirror those of the encoder, where upsampling is performed by nearest-neighbor interpolation. The LTE module operated on $k=256$ frequency coefficients. The INR network was a 1-conv residual network with 256 internal channels, 3 residual blocks, 2 subunits per block, with instance normalization between each layer. The SiLU activation function was used, following \cite{Pinaya.etal.2022a}. The Conv. LDM architecture also followed \cite{Pinaya.etal.2022a}, including the structure of the convolutional network, and the objective function including terms for an $\ell_1$ loss, a perceptual loss, an adversarial loss, and a KL latent space regularization term. 

\textbf{Experiment 2.} MedIL's convolutional backbone contained 3 residual blocks, kernel size 3, with 2 residual units per block, and 2 subunits per residual unit. Channel sizes were 32, 64, 96, and instance norm was used. Downscaling factors were 2, 2, with convolutional downsampling and upsampling being the same as in Experiment 1. The LTE module had $k=192$ frequency coefficients. The INR was a residual network with 128 internal channels, 2 residual blocks, and 2 subunits per block. The SiLU activation function was also used. 

The Conv. LDM architecture also followed \cite{Khader.etal.2023b}. For the objective function, we kept the following terms from the previous work: $\ell_1$ loss, adversarial loss, perceptual loss, and a KL regularization term. We did not include the codebook loss or the feature matching loss, but found it was not necessary as there was very high reconstruction performance from the Conv. LDM in Experiment 2.

\subsection{Training Details}
\label{subsec:appendix_train_details}

\textbf{Experiment 1.} MedIL was pre-trained on AC-PC aligned, 1mm isotropic data for 150 epochs. Then, both models were trained for 250 epochs with random patches of size $152 \times 184 \times 152$ on native-resolution data; an adversarial loss was used after epoch 125. Finally, the model decoder was fine-tuned for 50 epochs. MedIL used the AdamW \cite{Loshchilov.Hutter.2019} optimizer with a base learning rate of $0.005$, momentum of $0.9$, $\beta_2 = 0.999$, and weight decay of $0.01$. The Conv. LDM was trained for 300 epochs with an adversarial network used after 150 epochs. Checkpoints were selected by the $\ell_1$ loss of a validation set of 8 subjects.

\textbf{Training Time}. Training times for each T1w autoencoder model are given below. Note that due to the time required to load and preprocess large volumetric data files, training time may vary according to the number of available CPU cores.
\begin{itemize}
    \item Conv. LDM: 34 hours total on 4xA100 GPUs with 80 GB of GPU memory. 

    \item MedIL: 22 hours for pretraining, 25 hours for full autoencoder training, and an additional 2 hours for decoder-only training for a total of 49 hours. MedIL was trained on 4xA100 GPUs with 80 GB of GPU memory.

    \item MedIL-Interp: 26 hours for pretraining, 29 hours for full autoencoder training, and an additional 4 hours for decoder-only training for a total of 59 hours. MedIL-Interp was trained on 4xA6000 with 48 GB of GPU memory.
\end{itemize}

\textbf{Experiment 2.} Conv. LDM was trained for 250 epochs, with an adversarial term used after 25 epochs. MedIL and MedIL-Interp were pretrained on the $2.5$mm data for 150 epochs, then trained on native-space data for 125 epochs with no adversarial loss. The latent space had a resolution of $10$mm isotropic, with only scaling and translation between the input $X$ and latent volume $Z$. During MedIL's training stage, volumes were randomly augmented with a $20\%$ chance of resampling the input volume to a resolution between the native resolution and $0.8\times0.8\times2.5$mm. Gaussian noise with a random $\sigma \in [1.0, 70.0]$ was also added during augmentation.

\textbf{Training Time}. Training times for each T1w autoencoder model are given below. Note that due to the time required to load and preprocess large volumetric data files, training time may vary according to the number of available CPU cores.
\begin{itemize}
    \item Conv. LDM: 34 hours total on 4xA6000 GPUs with 48 GB of GPU memory. 

    \item MedIL: 12 hours for pretraining and 29 hours for full autoencoder training for a total of 41 hours. MedIL was trained on 4xA6000 GPUs with 48 GB of GPU memory.

    \item MedIL-Interp: 12 hours for pretraining and 26 hours for full autoencoder training for a total of 38 hours. MedIL-Interp was trained on 4xA6000 with 48 GB of GPU memory.
\end{itemize}

\section{Additional Reconstruction Results}

\begin{table}[]
\label{table:fixed_size_recon}
\caption{Reconstruction performance of MedIL from the pretraining stages in Experiments 1 and 2. Here, MedIL was trained on only fixed-size volumes, but evaluated on volumes in their native space.}

\centering
\begin{tabular}{l|lllll}
 &
  \cellcolor[HTML]{EFEFEF}$\downarrow$ $\ell_1$ &
  $\uparrow$PSNR &
  \cellcolor[HTML]{EFEFEF}$\uparrow$SSIM &
  $\uparrow$MS-SSIM &
  \cellcolor[HTML]{EFEFEF}$\downarrow$Perceptual \\ \hline
\textit{Experiment 1: Brain MRI} &                                  &           &                                  &          &                                  \\ \hline
MedIL Fixed Size                 & \cellcolor[HTML]{EFEFEF}0.046602 & 28.447635 & \cellcolor[HTML]{EFEFEF}0.831263 & 0.943564 & \cellcolor[HTML]{EFEFEF}0.000427 \\ \hline
\textit{Experiment 2: Lung CT}   &                                  &           &                                  &          &                                  \\ \hline
MedIL Fixed Size                 & \cellcolor[HTML]{EFEFEF}0.157682 & 20.348748 & \cellcolor[HTML]{EFEFEF}0.613703 & 0.741425 & \cellcolor[HTML]{EFEFEF}0.057822
\end{tabular}

\end{table}

We sought to understand the effect on MedIL's performance of training on native-space data vs. fixed-size data (similar to the fully convolutional autoencoder). So, for Experiments 1 and 2, we took the MedIL weights from the pretraining stage, which were only trained on fixed-size volumes, and had MedIL reconstruct volumes in the native space. The results are shown in Table \ref{table:fixed_size_recon}.







\end{document}